\documentclass{article}
\usepackage{spconf,amsmath,graphicx,subfigure}
\usepackage{booktabs} 

\def\onedot{. }
 
\def\ie{\emph{i.e}\onedot}

\def\wrt{w.r.t\onedot}

\usepackage{algorithm}
\usepackage[noend]{algpseudocode}
\makeatletter
\def\BState{\State\hskip-\ALG@thistlm}
\makeatother

\usepackage{epsfig}
\usepackage{amsmath}
\usepackage{amssymb}
\usepackage{booktabs}

\title{Context Propagation from Proposals for Semantic Video Object Segmentation}

\name{Tinghuai Wang}
\address{Nokia Technologies, Finland}

\begin{document}
%
\maketitle
\begin{abstract}

In this paper, we propose a novel approach to learning semantic contextual relationships in videos for semantic object segmentation. 
Our algorithm derives the semantic contexts from video object proposals which encode the key evolution of objects and the relationship
among objects over the spatio-temporal domain. This semantic contexts are propagated across the video to estimate the pairwise
contexts between all pairs of local superpixels which are integrated into a conditional random field in the form of pairwise potentials 
and infers the per-superpixel semantic labels. The experiments demonstrate that our contexts learning and propagation model effectively 
improves the robustness of resolving visual ambiguities in semantic video object segmentation compared with the state-of-the-art methods.

\end{abstract}
\section{Introduction}
\label{sec:intro}

Semantic video object segmentation, which aims to assign a semantic label to every pixel in video frames, is an essential step in various computer vision and multimedia analysis tasks. Recent years have witnessed significant attention and progress toward this problem. However, in addition to fast motion, appearance variations, pose change, and occlusions, the difficulty in resolving the inherent semantic ambiguities still plagues the robustness and accuracy of such a large scale semantic labeling problem. 

Recently, weakly supervised methods utilizing robust detection and tracking approach have been proposed to address the semantic ambiguity issue. In the learning process, object-relevant instances \ie object proposals sharing the same semantic tags are usually selected from video frames to learn category or instance specific models for semantic labeling. Early work trained classifiers to incorporate scene topology and semantics into pixel-level object detection and localization \cite{taylor2013semantic}. Both object detector and tracker were employed to either impose spatio-temporal coherence \cite{ZhangCLWX15,drayer2016object} or learn an appearance model \cite{wang2016semi} for encoding the appearance variation of semantic objects. Lately, video object proposal selection algorithm has also been proposed to integrate longer-range object reasoning with superpixel labeling \cite{wang2017a}. Despite of significant advances that have been made by the above methods, global contextual relationships between semantic video objects remain under-explored. Yet, contextual relationships are ubiquitous and provide important
cues for scene understanding related tasks. 

Pairwise contextual relationships have been investigated in image segmentation  \cite{lin2016efficient}  and object detection \cite{gidaris2015object,bai2017multiple} tasks, whose importance has been highlighted in effectively resolving semantic ambiguities.  However, these methods model the contextual relationships in terms of co-occurrence of higher-level statistics of object categories, which favors frequently appeared classes in the training data to enforce rigid semantic labelling. Small objects are more likely to be omitted due to the sensitivity to the number of pixels that objects occupy.

Graphical models have become powerful tools in computer vision \cite{wang2010multi,qi20173d,wang2017submodular,wang2019zero,wang2019graph,yang2021learning}, providing a versatile framework for modeling contextual relationships. These approaches exploit the inherent structure and relationships within images \cite{wang2015robust,tinghuai2016method,xing2021learning} and videos \cite{wang2010video,wang2014wide,chen2020fine}, facilitating more sophisticated and context-aware analysis. By representing visual elements as nodes and their interactions as edges, graphical models capture spatial, temporal, and semantic dependencies essential for tasks such as visual information retrieval \cite{hu2013markov}, stylization \cite{WangCSCG10,wang2011stylized,wang2013learnable}, object detection \cite{WangW16,tinghuai2016apparatus,zhao2021graphfpn}, scene understanding \cite{WangW14,wang2017cross,wang2020spectral,tinghuai2020watermark,deng2021generative}, and image or video segmentation \cite{wang2015weakly,wang2016semi,wang2016primary,tinghuai2017method,tinghuai2018method1,ZhuWAK19,zhu2019cross,tinghuai2020semantic,lu2020video,wang2021end}.

In this work, we propose a novel model to exploit and propagate contextual relationships among video object proposals without relying on training data. Such a way of modeling spatio-temporal object contextual relationships has not been well studied. Our model is able to capture the intra- and inter- category contextual relationships by considering the content of an input video in a non-parametric approach.  This context model is comprised of a set of spatial-temporal context exemplars which provide a novel interpretation of contextual relationships in a link view which formulates the problem of learning contextual relationships as the label propagation problem on a similarity graph. This similarity graph naturally reflects the intrinsic and extrinsic relationship between semantic objects in the spatial-temporal domain. Due to the sparsity of this similarity graph, the learning process can be very efficient.

\section{Our Approach}

In this section, we describe our proposed context model and how the learned contextual relationships are integrated into semantic labeling in a principled manner.

\subsection{Video Object Proposals}

We start by generating a set of video object proposals \wrt to semantic categories via object detection and temporal association which encode the long-range spatio-temporal evolution of various object features. Video object proposals are commonly tied to higher-level contexts such as object interactions and behaviours \cite{WangC12,WangW14,WangHC14,liu2015multiclass,wang2016primary,wang2017a,tang2018object}. 

Specifically, we utilize \cite{EndresH10} to extract generic object proposals from each frame of the input video, and run fast R-CNN \cite{girshick2015fast} on this pool of object proposals to detect objects \wrt the given semantic label. We keep a set of object hypotheses  $\mathbb{D}$ by removing the proposals with detection confidence lower than a threshold (0.5 in our system).
Similar to  \cite{wang2017a}, temporal association is applied on this cohort of object hypotheses $\mathbb{D}$ to generate tracks of video object proposals $\mathbb{T}$. Specifically, we utilize object tracker \cite{ma2015hierarchical} to track object hypotheses over time to both ends of the video sequence as follows. We firstly rank all remaining object hypotheses in $\mathbb{D}$ based on detection confidence; tracking is performed to both directions initialized by the bounding box of the highest ranked object hypothesis; object hypothesis in the new frame is selected and added to $T_i$ if it has a sufficient overlap, \ie Intersection-over-Union (IoU) higher than a threshold (0.5), with the tracker box; this object hypothesis is consequently removed from $\mathbb{D}$. This process is iteratively performed until no new trajectory hypothesis containing three or more object instances can be generated from $\mathbb{D}$. Superpixels \cite{felzenszwalb2004efficient} are extracted from each frame as the atomic data units. Let $\mathcal{R}_D$ be the set of superpixels constituting video object hypotheses, and $\mathcal{R}_U$ be the unlabeled superpixels. The parameters of \cite{felzenszwalb2004efficient} are set to produce fine superpixels in order to preserve object boundaries. 

\subsection{Graph Construction}

We derive our context model by initializing a k-nearest neighbor similarity graph $\mathcal{G}=(\mathcal{V},\mathcal{E})$ between all $N$ superpixels from  $\mathcal{R}_D \cup \mathcal{R}_U$. Each node $v_{i} \in \mathcal{V}$ of the graph is described by the L2-normalized \emph{fc6} feature $f_i$  of VGG-16 Net \cite{vggnet} in a forward-pass of the corresponding superpixel. The weight  $w_{i,j} \in \mathbf{W}$ of edge $e_{i,j} \in \mathcal{E}$ is computed as the inner-product between the feature vectors of neighboring nodes, i.e., $w_{i,j} = <f_i, f_j>$.

\subsection{Context Modeling}

We start the context modeling process by generating context exemplars. We consider frames which contain video object proposals as weak annotations, since the proposals normally capture essential parts of video objects. Let $\mathcal{F}$ be this set of annotated frames and $\mathcal{\hat{F}}$ be all the other frames in current video sequence. A context exemplar consists of a pair of superpixels and the corresponding semantic labels. The intuition behind this setting is that one superpixel with its semantic label supports the paired superpixel to be labeled with its corresponding semantic label. This exemplar is able to encode the global interaction and co-occurrence of semantic objects beyond local spatial adjacencies. The goal is to impose the consistency between each pair of superpixels from un-annotated frames and the extracted context exemplars. 

Formally, given a set of semantic labels 
\begin{equation} 
\mathcal{L} = \{l_0, l_1, \dots, l_{C-1}\}  \nonumber
\end{equation}
comprising all $C$ classes in the annotated frame, we represent the context exemplars for each class pair $(c_m, c_n)$ as
\begin{equation} 
\mathbf{Q^{m,n}}  = \{(v_i, v_j): L(v_i) = l_m, L(v_j) = l_n, v_i, v_j \in \mathcal{F} \} \nonumber
\end{equation}
where $v_i, v_j \in \mathcal{F}$ stands for two superpixels $v_i$ and $v_j$ from the annotated frame set $\mathcal{F}$ and 
$L(v_i)$ represents the semantic label of superpixel $v_i$. Hence, all object class pairs as well as contextual relationships in the annotated frames are represented as 
\begin{equation} 
\mathcal{Q} = \{\mathbf{Q}^{0,0}, \mathbf{Q}^{0,1}, \dots, \mathbf{Q}^{C-1,C-1}\}. \nonumber
\end{equation}

We transform the above context exemplar to a context link view, where context exemplar $(v_i, v_j)$ can be referred to as a $(l_m, l_n)$-type link between two nodes on the similarity graph. Let $\mathcal{H}$ denote the set of $N\times N$ matrices, where a matrix $\mathbf{H}^{m,n}\in \mathcal{H}$ is associated with all $(l_m, l_n)$ class pair links.  Each entry $[\mathbf{H}^{m,n}]_{i,j} \in \mathbf{H}^{m,n}$ indicates the confidence of $(l_m, l_n)$-link between two superpixels $v_i$ and $v_j$. The confidence indicates the probability of the existence of a link.
The $(l_m, l_n)$-links which have been observed within the annotated frames can be represented by another set of matrices $\mathbf{P}^{m,n}\in \mathcal{H}$ such that
\begin{equation} 
[\mathbf{P}^{m,n}]_{i,j} =
\left\{ \begin{array}{lll}
1 &  \mbox{if} & (v_i, v_j) \in \mathbf{Q^{m,n}}  \\ 
0 & \mbox{otherwise} &
\end{array}\right.
\end{equation}
All the observed context link can be denoted as 
\begin{equation} 
\mathcal{P} = \{\mathbf{P}^{0,0}, \mathbf{P}^{0,1}, \dots, \mathbf{P}^{C-1,C-1}\}. \nonumber
\end{equation}

\subsection{Context Propagation}

We formulate the context prediction problem as a link prediction problem which infers how probable a certain link exists in a graph.
Specifically, we predict $(l_m, l_n)$-links among the pairs of nodes from $\mathcal{R}_U$ based on  $\mathbf{P}^{m,n}$ consistent to the intrinsic structure of the similarity
graph. To this end, we propagate $(l_m, l_n)$-links in $\mathbf{P}^{m,n}$ to estimate the strength of the pairs of nodes from $\mathcal{R}_U$. We drop the $m,n$
suffix for clarity. 

\begin{algorithm}
	\caption{Context learning algorithm}\label{algo}
	\begin{algorithmic}[1]
		\Procedure{Link prediction}{}
		\State $S (v_i, l_i, v_j, l_j) \gets \varnothing$
		\State $\text{Graph} ~\mathcal{G} \gets \text{all}~\textit{superpixels of video}$
		\State $\text{Affinity matrix} ~\mathbf{W} \gets \textit{k-nearest neighbors} $
		\State $d_i=\sum_{j=1}^{N} w_{ij}$
		\State $\mathbf{D} \gets \mathrm{diag}([d_1, \dots, d_N])$
		\State $\mathbf{L} \gets \mathbf{D}^{-\frac{1}{2}} \mathbf{W}  \mathbf{D}^{-\frac{1}{2}}$
		\State $\mathcal{Q} \gets \textit{context exemplars} $
		\State $\mathcal{P} \gets \textit{context links} \in \mathcal{Q}$
		\For{\text{each} \textit{class pair $(l_m, l_n)$}}
		\State $\mathbf{H}_r(1)  \gets \mathbf{0}, ~ \mathbf{H}_c(1)  \gets \mathbf{0}$
		\State $\textit{Convergence} \gets \mathbf{false}$
		\While{$\textit{Convergence} ~\text{is}~ \mathbf{false}$} \Comment{row-wise}
		\State $\mathbf{H}_r(t+1) \gets \mu \mathbf{L} \mathbf{H}_r(t) + (1-\mu)\mathbf{P}^{m,n}$ 
		\EndWhile\label{euclidendwhile}
		\State $\textit{Convergence} \gets \mathbf{false}$
		\While{$\textit{Convergence} ~\text{is}~ \mathbf{false}$} \Comment{column-wise}
		\State $\mathbf{H}_c(t+1) \gets \mu \mathbf{L} \mathbf{H}_c(t) + (1-\mu)\mathbf{\hat{H}}_r$ 
		\EndWhile\label{euclidendwhile}	
		\State $S (v_i, l_i = l_m, v_j, l_j = l_n) \gets [\mathbf{\hat{H}}_c]_{ij}$
		\EndFor
		\EndProcedure
	\end{algorithmic}
\end{algorithm}

It is  impractical to directly solve the  link prediction problem for video segmentation due to the prohibited complexity $O(N^4)$, and thus we 
propose to solve it by decomposing it into two separate label propagation processes. As described in Algorithm \ref{algo},
row-wise link predication (step 13-14) is firstly performed, followed by column-wise link prediction (step 16-17). Specifically, the $j$-th row $\mathbf{P}^{j,.}$,
\ie the context exemplars associated with $v_j$, serves as an initial configuration of a label propagation problem \cite{Zhou2004} with respect to node $v_j$.
Each row is handled separately as a binary label propagation which converges to $\mathbf{\hat{H}}_r$. It is observed that the label propagation does not
apply to the rows of $\mathbf{P}$ corresponding to $\mathcal{R}_U$, and thus we only perform row-wise link propagation in rows corresponding to
annotated superpixels, which is much less than $N$. For the column-wise propagation, the $i$-th converged row $[\mathbf{\hat{H}}_r]_i$ is used 
to initialize the configuration. After convergence of the column-wise propagation, the probability of $(l_m, l_n)$-link between two nodes
of $\mathcal{R}_U$ is obtained. 

\subsection{Semantic Labeling}

Semantic video object segmentation problem can be formulated as a superpixel labeling problem.
We now describe how to integrate the predicted contextual relationship into the semantic labeling process.  
Since we have learned context link scores $S (v_i, l_i, v_j, l_j)$, it is straight forward to incorporate them to 
 the fully connected CRF  which is proved to be effective in encoding model contextual relationships between object classes. 

We define a random field $\mathbf{x}$ defined over a set of variables $\{\mathrm{x}_0, \dots, \mathrm{x}_{N-1}\}$, and the domain of each variable is a set of class
labels $\mathcal{L} = \{l_0, l_1, \dots, l_{C-1}\}$. The corresponding Gibbs energy is 
\begin{equation} 
E(\mathbf{x}) = \sum_{i} \psi(\mathrm{x}_i) +  \sum_{i,j} \phi(\mathrm{x}_i, \mathrm{x}_j). \label{eq:graphcut}
\end{equation}

The pairwise potential $\phi(\mathrm{x}_i, \mathrm{x}_j)$ encodes the contextual relationships between the superpixels learned via link prediction, which is defined 
as
\begin{equation} 
\phi(\mathrm{x}_i, \mathrm{x}_j) = \exp (-\frac{S (v_i, c_i, v_j, c_j)^2}{2\beta}) 
\end{equation}
where $\beta = <S (v_i, l_i, v_j, l_j)^2>$ is the adaptive weight and $<\cdot>$ indicates the expectation. 

The unary potential $\psi(\mathrm{x}_i)$ is defined as the negative logarithm of the likelihood of assigning $v_i$ with label $\mathrm{x}_i$. 
To obtain $\psi(\mathrm{x}_i)$, we learn a SVM 
model based on hierarchical CNN features \cite{ma2015hierarchical} by sampling from the 
annotated frames.

We adopt $\alpha$-expansion  \cite{BoykovVZ01} to optimize (\ref{eq:graphcut})  
and the resulting label assignment gives the semantic object segmentation of the video sequence. 

\section{Experiments}
\label{sec:evaluation}

We evaluate our proposed approach on YouTube-Objects \cite{PrestLCSF12} which is the \emph{de facto} benchmark for assessing semantic video object segmentation algorithms. YouTube-Objects contains over 30,000 frames in total with dense annotations for 20 classes of PASCAL VOC 2012.  These videos are very challenging and completely unconstrained, with objects of similar colour to the background, fast motion, non-rigid deformations, and fast camera motion. Standard average IoU is used to measure the segmentation accuracy, $IoU = \frac{S \cap G}{S \cup G}$, where $S$ is the segmentation result and $G$ stands for the ground-truth mask. We compare our approach with $8$ state-of-the-art automatic approaches on this dataset, including two motion driven segmentation \cite{BroxM10,Papazoglou2013}, three weakly supervised semantic segmentation approaches \cite{PrestLCSF12,TangSY013,ZhangCLWX15}, and detection based approaches \cite{drayer2016object,wang2017a,wang2017}.

\begin{table}[t!]
	\caption{Intersection-over-union overlap accuracies on YouTube-Objects Dataset}
	\centering 
	\resizebox{\columnwidth}{!}{
	\begin{tabular}{l|rrrrrrrrr} 
		\toprule
		& \cite{BroxM10} & \cite{PrestLCSF12} & \cite{Papazoglou2013} & \cite{TangSY013}  &  \cite{ZhangCLWX15} & \cite{drayer2016object} &  \cite{wang2017a} &  \cite{wang2017}  & Ours\\
		\midrule
		Plane & 0.539 & 0.517  & 0.674  & 0.178  & 0.758    & 0.744   & 0.703 & 0.757 & \textbf{0.762}\\
		Bird    & 0.196 & 0.175  & 0.625  & 0.198  & 0.608   & 0.721   & 0.631 & 0.766 & \textbf{0.767}\\
		Boat   & 0.382 & 0.344  & 0.378  & 0.225  & 0.437  & 0.585     & 0.659 & 0.666 & \textbf{0.702} \\
		Car     & 0.378 & 0.347  & 0.670  & 0.383  &  0.711  & 0.600      & 0.625 &  \textbf{0.758}  &0.747\\
		Cat	  & 0.322 & 0.223  & 0.435  & 0.236  & 0.465    & 0.457   &   0.497 & 0.624 & \textbf{0.659}\\
		Cow   & 0.218 & 0.179  & 0.327   & 0.268  & 0.546   &  0.612     &  0.701 &  \textbf{0.720} &0.719\\
		Dog    & 0.270 & 0.135  & 0.489   & 0.237  & 0.555   & 0.552    & 0.532 & 0.671 & \textbf{0.669}\\
		Horse & 0.347 & 0.267  & 0.313   & 0.140  & 0.549     &  0.566    &  0.524 & 0.526 & \textbf{0.593}\\
		Mbike & 0.454 & 0.412  & 0.331   & 0.125  & 0.424    & 0.421 	&  0.554 & 0.547 & \textbf{0.605}\\
		Train   & 0.375 & 0.250  & 0.434    & 0.404  & 0.358	& 0.367  	&  0.411  & 0.392 & \textbf{0.411} \\
		\midrule
		\shortstack{Avg.}  &0.348 &0.285 &0.468   & 0.239  & 0.541    & 0.562  	& 0.584 & 0.643 & \textbf{0.664}\\
		\bottomrule
	\end{tabular} \label{tbl:yto-result} }
\end{table}

\begin{figure*}[t!]
	\centering
	\includegraphics[width=0.99\linewidth]{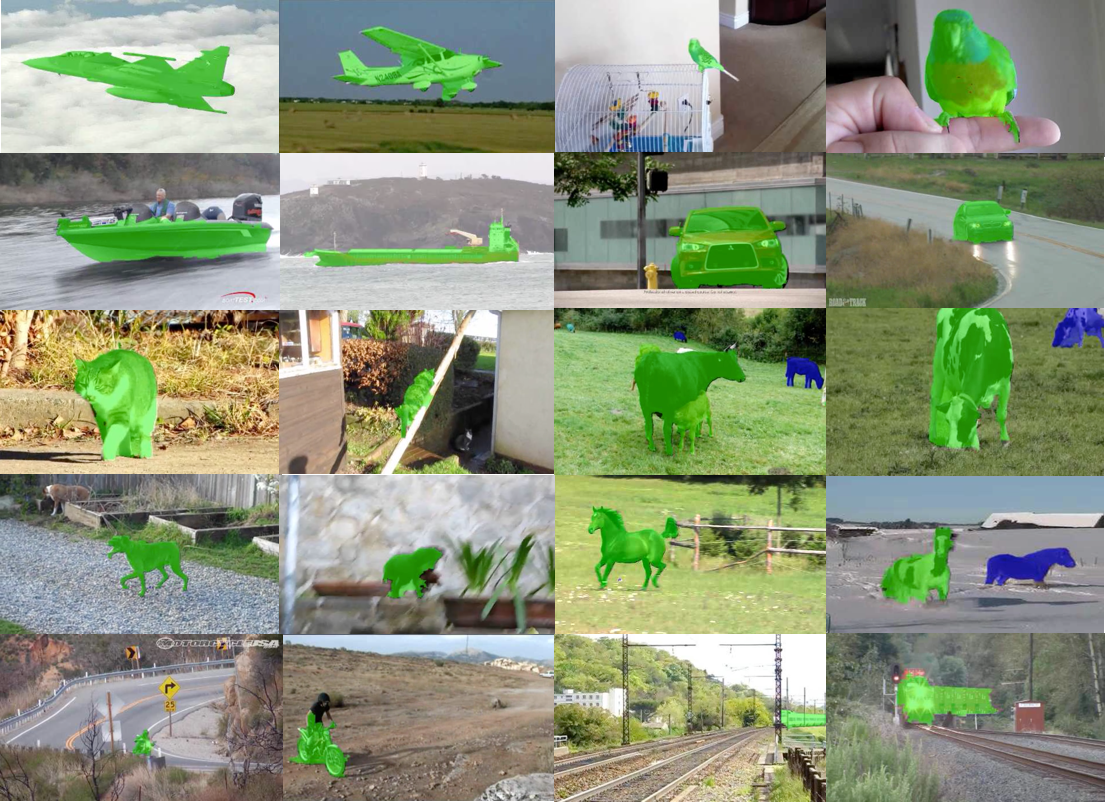}
	\caption{Qualitative results of our algorithm on YouTube-Objects Dataset.  \label{fig:yto}}
\end{figure*}

As summerized in Table \ref{tbl:yto-result}, our proposed algorithm outperforms the compared methods, with a considerable margin of $2.1\%$ in average over the best competing method \cite{wang2017}. In terms of category-wise comparison, our approach surpasses $8$ out of $10$ categories. This improvement over the state-of-the-art is owing to the capability of learning
and propagating higher-level spatial-temporal contextual relationships of video object proposals which capture the key semantic contexts in challenging video data, 
as opposed to imposing contextual information in local labeling (\cite{wang2017}) or modeling local appearance ( \cite{ZhangCLWX15,wang2017a}).  One common limitation of the these methods is that they are error-prone in separating interacting objects exhibiting similar appearance or motion, which is intractable unless the inherent contextual relationship is explored.

Our algorithm outperforms another two methods which also utilize object detection, \ie  \cite{drayer2016object}  and \cite{ZhangCLWX15}, with large margins of $10.2\%$ and $12.3\%$.  \cite{drayer2016object} uses R-CNN for the initial object detection and incorporates the detection scores in a graph cut optimization, which, however, does not explore the semantic contextual information. \cite{ZhangCLWX15} performs the worst since it only conducts temporal association of detected object segments without explicitly modeling either the objects or contexts.

The approaches which train discriminative classifiers given weakly annotated videos, \ie  \cite{Papazoglou2013} and \cite{TangSY013}, perform relatively worse than detection based approaches. The trained discriminative classifiers can be easily overfitted to the small amount of annotated data, which makes them difficult to tackle fast-changing natural scenes. As a contrast, our approach is able to exploit semantic contextual relationships to compensate the limited capabilities of object detection on object with unusual viewpoints. Some qualitative results of the proposed algorithm on YouTube-Objects dataset are shown in Fig. \ref{fig:yto}.

\section{Conclusion}

We have proposed a novel approach to modeling the semantic contextual relationships for tackling the challenging video object segmentation problem. The proposed model comprises an exemplar-based nonparametric view of contextual cues, which is formulated as link prediction problem solved by label propagation on a similarity graph of superpixels. The derived contextual relationships are utilized to estimate the pairwise contexts between all pairs of unlabeled local superpixels. The experiments demonstrated that modeling the semantic contextual relationships effectively improved
segmentation robustness and accuracy which significantly advanced the state-of-the-art on challenging benchmark.

\bibliographystyle{IEEEbib}
\bibliography{refs}

\end{document}